\documentclass{article}
\usepackage[pdftex]{graphicx}
\usepackage{amsmath,epsfig,bm, amssymb,subfigure}

\newcommand{\pix}{\pi^{\mathrm x}}
\newcommand{\piy}{\pi^{\mathrm y}}

\newcommand{\pxy}{p_{\mathrm{xy}}}
\newcommand{\px}{p_{\mathrm x}}
\newcommand{\py}{p_{\mathrm y}}

\newcommand{\argmax}{\mathop{\mathrm{argmax\,}}}

\newcommand{\mathbbR}{\mathbb{R}}

\newcommand{\boldE}{{\boldsymbol{E}}}

\newcommand{\boldH}{{\boldsymbol{H}}}
\newcommand{\boldI}{{\boldsymbol{I}}}

\newcommand{\boldM}{{\boldsymbol{M}}}

\newcommand{\boldU}{{\boldsymbol{U}}}
\newcommand{\boldV}{{\boldsymbol{V}}}
\newcommand{\boldW}{{\boldsymbol{W}}}
\newcommand{\boldX}{{\boldsymbol{X}}}
\newcommand{\boldY}{{\boldsymbol{Y}}}
\newcommand{\boldZ}{{\boldsymbol{Z}}}

\newcommand{\boldh}{{\boldsymbol{h}}}

\newcommand{\boldx}{{\boldsymbol{x}}}
\newcommand{\boldy}{{\boldsymbol{y}}}

\newcommand{\boldalpha}{{\boldsymbol{\alpha}}}

\newcommand{\boldpi}{{\boldsymbol{\pi}}}

\newcommand{\boldPi}{{\boldsymbol{\Pi}}}

\newcommand{\calF}{{\mathcal{F}}}

\newcommand{\calZ}{{\mathcal{Z}}}

\usepackage{url,multirow}

\advance\oddsidemargin-1.0in
\date{\today}
\title{Dependence Maximizing Temporal Alignment \\via Squared-Loss Mutual Information} 

\author{Makoto Yamada$^1$, Leonid Sigal$^2$, Michalis Raptis$^2$, and Masashi Sugiyama$^{1}$\\
$^1$Tokyo Institute of Technology, 2-12-1 O-okayama, Meguro-ku, Tokyo 152-8552, Japan\\
$^2$Disney Research Pittsburgh, 4720 Forbes Ave., Pittsburgh, PA 15213\\
\texttt{\{yamada@sg. \hspace{-3mm}sugi@\}cs.titech.ac.jp}}

\begin{document}
\maketitle

\begin{abstract}
The goal of \emph{temporal alignment} is to establish time correspondence
between two sequences,
which has many applications in a variety of areas
such as speech processing, bioinformatics, computer vision, and computer graphics.
In this paper, we propose a novel temporal alignment method
called \emph{least-squares dynamic time warping} (LSDTW).
LSDTW finds an alignment that maximizes statistical dependency between sequences,
measured by a squared-loss variant of mutual information.
The benefit of this novel information-theoretic formulation is that
LSDTW can align sequences with different lengths, different dimensionality,
high non-linearity, and non-Gaussianity
in a computationally efficient manner.
In addition, model parameters such as an initial alignment matrix
can be systematically optimized by cross-validation.
We demonstrate the usefulness of LSDTW through experiments on synthetic
and real-world \emph{Kinect} action recognition datasets.
\end{abstract}

\section{Introduction}
Temporal alignment of sequences is an important problem with many practical applications such as
speech recognition \cite{saoke78,Rabiner:1993dq}, activity recognition \cite{DBLP:journals/pami/JunejoDLP11,DBLP:conf/iccv/GongM11}, temporal segmentation \cite{DBLP:conf/fgr/ZhouTH08}, curve matching \cite{Sebastian03b.b.:on}, chromatographic and micro-array data analysis \cite{Listgarten05multiplealignment}, synthesis of human motion \cite{Hsu05styletranslation}, and temporal alignment of human motion \cite{NIPS2009_0760,CVPR:Feng:2012a}. %

\emph{Dynamic time warping} (DTW) is a classical temporal alignment method that aligns two sequences by minimizing the pairwise squared Euclidean distance \cite{saoke78,Rabiner:1993dq}. An advantage of DTW is that the minimization can be efficiently carried out by \emph{dynamic programming} (DP) \cite{bellman52}. 
However, due to the Euclidean formulation, DTW may not be able to find a good alignment when the characteristics of the two sequences are substantially different (e.g., sequences have different amplitudes). 
Moreover, DTW cannot handle sequences with different dimensions (e.g., image to audio alignment),
which limits the range of applications significantly.

To overcome the weaknesses of DTW, \emph{canonical time warping} (CTW) was introduced \cite{NIPS2009_0760}. 
CTW performs sequence alignment in a common latent space found by 
canonical correlation analysis (CCA) \cite{hotelling36cca}.
Thus, CTW can naturally handle sequences with different dimensions.
However, CTW can only deal with linear projections,
and it is difficult to optimize
model parameters such as the initial alignment matrix, the regularization parameter used in CCA, and the dimensionality of the common latent space.

To handle non-linearity, \emph{dynamic manifold temporal warping} (DMTW) was recently proposed in \cite{DBLP:conf/iccv/GongM11}. DMTW first transforms original data onto a one-dimensional non-linear manifold and then finds an alignment on this manifold using DTW.
Although DMTW is highly flexible by construction, its performance depends heavily on the choice of the non-linear transformation and, moreover, it implicitly assumes the smoothness of sequences. 
For this reason, DMTW has limited applicability.

In this paper,
we propose a novel information-theoretic
temporal alignment method based on statistical dependence maximization.
Our method, which we call \emph{least-squares dynamic time warping} (LSDTW), employs
a squared-loss variant of mutual information called
\emph{squared-loss mutual information} (SMI) as a dependency measure.
SMI is estimated by the method of
\emph{least-squares mutual information} (LSMI) \cite{BMCBio:Suzuki+etal:2009a},
which is consistent estimator achieving the optimal non-parametric convergence rate to the true SMI.
An advantage of the proposed LSDTW over existing methods is that
it can naturally deal with non-linearity and non-Gaussianity in data through SMI.
Moreover, cross-validation (CV) with respect to the LSMI criterion is possible,
which allows selection of model parameters such as the initial alignment matrix, the Gaussian kernel width, and the regularization parameter.
Furthermore, the formulation of LSDTW is quite general
and does not require strong assumptions on the topology of the latent manifold (e.g., smoothness).
Thus, LSDTW is expected to perform well in a broader range of applications.
Indeed, through experiments on synthetic and real-world \emph{Kinect} action recognition tasks,
LSDTW is shown to be a promising alternative to existing temporal alignment methods.

\section{Dependence Maximizing Temporal Alignment via SMI}
In this section, we first formulate the problem of \emph{dependence maximizing temporal alignment} (DMTA) and then develop a DMTA method based on \emph{squared-loss mutual information} (SMI)  \cite{BMCBio:Suzuki+etal:2009a}.
\subsection{Formulation of Dependence Maximizing Temporal Alignment (DMTA)} \label{sec:prob_formulate}
\label{sec:Problem}


Given two sequences represented by a set of  samples (ordered in time), 
\[
\{\boldx_i\;|\; \boldx_i \in \mathbbR^{d_{\mathrm x}}\}_{i=1}^{n_{\mathrm x}}
\mbox{ and }
\{\boldy_j\;|\; \boldy_j \in \mathbbR^{d_{\mathrm y}}\}_{j=1}^{n_{\mathrm y}},
\]
the goal of DMTA is to find a temporal alignment such that the statistical dependency between two sets of samples is maximized. Note that $n_{\mathrm x}$ and $d_{\mathrm x}$ can, in general, be different from $n_{\mathrm y}$ and $d_{\mathrm x}$.

Let $\boldpi^{\mathrm x}$ and $\boldpi^{\mathrm y}$ be alignment functions over $\{1,\ldots,n_{\mathrm x}\}$ and $\{1,\ldots,n_{\mathrm y}\}$,
and let $\boldPi$ be the corresponding alignment matrix:
\begin{align*}
\boldPi &:= [\boldpi^{\mathrm x}~\boldpi^{\mathrm y}]^{\top} \in  \mathbbR^{2 \times m},\\
\boldpi^{\mathrm x}&:= [\pix_1, \ldots, \pix_{m}]^\top \in \{1,\ldots,n_{\mathrm x}\}^{m \times 1}, \\
\boldpi^{\mathrm y}&:= [\piy_1, \ldots, \piy_{m}]^\top \in \{1,\ldots,n_{\mathrm y}\}^{m \times 1},
\end{align*}
where $m$ is the number of indexes needed to align the sequences and $^\top$ denotes the transpose. $\boldPi$ needs to satisfy the following three additional constraints:\\
\noindent{\bf$\bullet$ Boundary condition:} $[\pix_1~ \piy_1]^\top = [1~1]^\top$ and $[\pix_{m}~ \piy_{m}]^\top = [n_{\mathrm x}~n_{\mathrm y}]^\top$.\\
\noindent{\bf$\bullet$ Continuity condition:} $0 \leq \pix_t - \pix_{t-1}  \leq 1$ and $0 \leq \piy_t - \piy_{t-1} \leq 1$.\\
\noindent{\bf$\bullet$ Monotonicity condition:} $t_1 \geq t_2 \rightarrow \pix_{t_1} \geq \pix_{t_2} , \piy_{t_1} \geq \piy_{t_2}$.

Let us denote the paired samples aligned by $\boldpi^{\mathrm x}$ and $\boldpi^{\mathrm y}$ as
\[
Z(\boldPi) := \{(\boldx_{\pix_i},\boldy_{\piy_i})\}_{i=1}^m.
\]
Then, the optimal alignment, denoted by $\boldPi^\ast$, is defined as the maximum of a certain statistical dependence measure $D$ between the two sets $\{\boldx_i\}_{i=1}^{n_{\mathrm x}}$ and $\{\boldy_j\}_{j=1}^{n_{\mathrm y}}$:
\begin{align*}
\boldPi^* := \mathop{\text{argmax}}_{\boldPi} D(Z(\boldPi)).
\end{align*}

\subsection{Least-Squares Dynamic Time Warping (LSDTW)}
A popular measure of statistical dependence is \emph{mutual information} (MI) \cite{Bell:Shannon:1948},
and its estimation has been studied thoroughly
\cite{IEEE-IT:Wang+etal:2005,AISM:Sugiyama+etal:2008,ISIT:Perez-Cruz:2008,JSPI:Silva+Narayanan:2010,IEEE-IT:Nguyen+etal:2010}.
However, these MI approximations are computationally expensive,
mainly due to the non-linearity introduced by the ``log'' function.
In this paper, we propose to use a squared-loss variant of MI called
\emph{squared-loss MI} (SMI), which results in a simple and computationally efficient estimation algorithm
called \emph{least-squares dynamic time warping} (LSDTW).

\subsubsection{Overview}
The optimization problem of LSDTW is defined as
\begin{align}
\boldPi^* := \mathop{\text{argmax}}_{\boldPi} \textnormal{SMI}(Z(\boldPi)),
\label{eq:LSDTW-Optimization-Problem}
\end{align}
where SMI is defined and expressed as
\begin{align}
\textnormal{SMI}(Z)  &= \frac{1}{2}\iint \left(\frac{\pxy(\boldx,\boldy)}{\px(\boldx)\py(\boldy)} - 1\right)^2 \px(\boldx) \py(\boldy) \textnormal{d}\boldx \textnormal{d}\boldy,
\nonumber\\
&= \frac{1}{2}\iint \left(\frac{\pxy(\boldx,\boldy)}{\px(\boldx)\py(\boldy)}\right)\pxy(\boldx,\boldy)\textnormal{d}\boldx \textnormal{d}\boldy - \frac{1}{2},
\label{eq:SMIdef}
\end{align}
where
$\pxy(\boldx,\boldy)$ is the joint density of $\boldx$ and $\boldy$,
and $\px(\boldx)$ and $\py(\boldy)$ are the marginal densities of $\boldx$ and $\boldy$,
respectively.
SMI is the \emph{Pearson divergence}
\cite{PhMag:Pearson:1900} from $\pxy(\boldx,\boldy)$ to $\px(\boldx)\py(\boldy)$,
while the ordinary MI is the \emph{Kullback-Leibler divergence}
\cite{Annals-Math-Stat:Kullback+Leibler:1951} from $\pxy(\boldx,\boldy)$ to $\px(\boldx)\py(\boldy)$.
SMI is non-negative and is zero if and only if $\boldx$ and $\boldy$ are statistically independent, as the ordinary MI. 

Based on Eq.\eqref{eq:LSDTW-Optimization-Problem}, we develop the following iterative algorithm for estimating $\boldPi$:
\begin{description}
\item[(i)  Initialization:] Initialize the alignment matrix $\boldPi$.
\item[(ii) Dependence estimation:] For the current $\boldPi$, obtain an SMI estimator $\widehat{\textnormal{SMI}}(Z(\boldPi))$.
\item[(iii) Dependence maximization:] Given an SMI estimator $\widehat{\textnormal{SMI}}(Z(\boldPi))$, obtain the maximum alignment $\boldPi$.
\item[(iv) Convergence check:] The above (ii) and (iii) are repeated until $\boldPi$ fulfills a convergence criterion.
\end{description}
\subsubsection{Dependence Estimation}
\label{subsec:dependence-estimation}
In the dependence estimation step, we utilize a non-parametric SMI estimator called
\emph{least-squares mutual information} (LSMI) \cite{BMCBio:Suzuki+etal:2009a},
which was shown to possess a superior convergence property
\cite{AISTATS:Suzuki+Sugiyama:2010}.
Here, we briefly review LSMI.

{\bf Basic Idea:} The key idea of LSMI is to directly estimate the \emph{density ratio} in Eq.\eqref{eq:SMIdef}
\cite{book:Sugiyama+etal:2012},
\[
r(\boldx, \boldy) := \frac{\pxy(\boldx, \boldy)}{\px(\boldx)\py(\boldy)},
\]
 from paired samples $Z(\boldPi) = \{(\boldx_{\pix_i},\boldy_{\piy_i})\}_{i=1}^m$ without going through density estimation of $\pxy(\boldx, \boldy)$, 
$\px(\boldx)$, and $\py(\boldy)$.
Here, the density-ratio function $r(\boldx, \boldy)$ is
directly modeled as
\begin{eqnarray}
r_{\boldalpha}(\boldx, \boldy)= \sum_{\ell = 1}^m \alpha_{\ell} K(\boldx,\boldx_{\pix_\ell})L(\boldy,\boldy_{\piy_{\ell}}),
\label{ratio-model}
\end{eqnarray}
where $K(\boldx,\boldx')$ and $L(\boldy,\boldy')$ are kernel functions (e.g., Gaussian kernels) for $\boldx$
and $\boldy$, respectively.

Then, the parameter $\boldalpha=(\alpha_1,\ldots,\alpha_m)^\top$ is learned so that the squared error to the true density ratio is minimized:
\begin{align*}
J_0(\boldalpha):=\frac{1}{2}\iint (r_{\boldalpha}(\boldx,\boldy) - r(\boldx,\boldy))^2 \px(\boldx) \py(\boldy) \textnormal{d}\boldx \textnormal{d}\boldy.
\end{align*}
After a few lines of calculation, $J_0$ can be expressed as
\begin{align*}
J_0(\boldalpha) = J(\boldalpha)+\textnormal{SMI}(Z(\boldPi))+\frac{1}{2},
\end{align*}
where
\begin{align*}
 J(\boldalpha) &:=
\frac{1}{2}\boldalpha^\top \boldH_{\boldPi} \boldalpha -\boldh_{\boldPi}^\top\boldalpha, \\
H_{\boldPi,\ell,\ell'} &:=  \iint K(\boldx,\boldx_{\pix_\ell})L(\boldy,\boldy_{\piy_{\ell}}) K(\boldx,\boldx_{\pix_{\ell'}})L(\boldy,\boldy_{\piy_{\ell'}}) \px(\boldx) \py(\boldy) \textnormal{d}\boldx \textnormal{d}\boldy, \\
h_{\boldPi,\ell} &:= \iint K(\boldx,\boldx_{\pix_\ell})L(\boldy,\boldy_{\piy_{\ell}}) \pxy(\boldx,\boldy)\textnormal{d}\boldx \textnormal{d}\boldy.
\end{align*}
Since $\textnormal{SMI}(Z(\boldPi))$ is constant with respect to $\boldalpha$,
minimizing $J_0$ is equivalent to minimizing $J$.

{\bf Computing the Solution:} Approximating the expectations in $\boldH_\boldPi$ and $\boldh_\boldPi$ included in $J$
by empirical averages,
we have the following optimization problem:
\begin{align}
\min_{\boldalpha} \left[
  \frac{1}{2}\boldalpha^\top \widehat{\boldH}_\boldPi\boldalpha
  -\widehat{\boldh}_\boldPi^\top\boldalpha
 + \frac{\lambda}{2} \boldalpha^\top\boldalpha \right], 
\label{eq:SMI_cost_func}
\end{align}
where $\lambda\boldalpha^\top \boldalpha/2$ is the regularization term to avoid overfitting, $\lambda$ ($\ge0$) is the regularization parameter, and
\begin{align*}
\widehat{H}_{\boldPi,\ell,\ell'} &:= \frac{1}{m^2} 
\sum_{i,j = 1}^mK(\boldx_{\pix_i},\boldx_{\pix_\ell})L(\boldy_{\piy_j},\boldy_{\piy_{\ell}}) K(\boldx_{\pix_i},\boldx_{\pix_{\ell'}})L(\boldy_{\piy_j},\boldy_{\piy(\ell')}),\\
\widehat{h}_{\boldPi,\ell} &:= \frac{1}{m}\sum_{i = 1}^m 
K(\boldx_{\pix_i},\boldx_{\pix_\ell})L(\boldy_{\piy_i},\boldy_{\piy_{\ell}}).
\end{align*}
Differentiating Eq.\eqref{eq:SMI_cost_func} with respect to $\boldalpha$ and equating it to zero,
we can obtain the optimal solution $\widehat{\boldalpha}_\boldPi$ analytically as
\begin{align}
\label{eq:LSMI_Solution}
\widehat{\boldalpha}_\boldPi= (\widehat{\boldH}_\boldPi + \lambda \boldI)^{-1}\widehat{\boldh}_\boldPi,
\end{align}
where $\boldI$ is the $m \times m$ identity matrix. Note that, LSMI has time complexity $O(m^3)$ due to the matrix inversion. However, when the number of training data is large, we can reduce the number of kernels in Eq.\eqref{ratio-model} to $l (<m)$ by sub-sampling. With this approximation, the inverse matrix in Eq.\eqref{eq:LSMI_Solution} can be computed with time complexity $O(l^3)$.

Finally, the following SMI estimator can be obtained by taking the empirical average of Eq.\eqref{eq:SMIdef} as
\begin{align}
\widehat{\mathrm{SMI}}(Z(\boldPi))&= \frac{1}{2m} \sum_{i = 1}^m r_{\widehat{\boldalpha}_\boldPi}(\boldx_{\pix_i}, \boldy_{\piy_i}) - \frac{1}{2}.
\label{SMIhat}
\end{align}

{\bf Model Selection:} \label{subsubsec:LSMI-CV} Hyper-parameters included in the kernel functions
and the regularization parameter can be optimized by cross-validation 
with respect to $J$ \cite{BMCBio:Suzuki+etal:2009a},
which is described below.

First, samples $\calZ=\{(\boldx_i, \boldy_i)\}_{i=1}^n$ are
divided into $K$ disjoint subsets $\{\calZ_k\}_{k=1}^K$ of (approximately) the same size.
Then, an estimator $\widehat{\boldalpha}_{\calZ_k}$ is obtained using $\calZ\backslash\calZ_k$
(i.e., all samples without $\calZ_k$),
and the approximation error for hold-out samples $\calZ_k$ is computed as 
\[
J_{\calZ_k}^{(K \text{-CV})} := \frac{1}{2}\widehat{\boldalpha}_{\calZ_k}^\top \widehat{\boldH}_{\calZ_k}\widehat{\boldalpha}_{\calZ_k} - \widehat{\boldh}_{\calZ_k}^\top\widehat{\boldalpha}_{\calZ_k},
\]
where,
for $|\calZ_k|$ being the number of samples in subset $\calZ_k$, 
\begin{align*}
[\widehat{H}_{\calZ_k}]_{\ell,\ell'} &:= \frac{1}{|\calZ_k|^2} 
\sum_{\boldx\in\calZ_k}
\sum_{\boldy\in\calZ_k}
K(\boldx,\boldx_\ell)L(\boldy,\boldy_\ell) K(\boldx,\boldx_{\ell'})L(\boldy,\boldy_{\ell'}),\\
[\widehat{h}_{\calZ_k}]_\ell &:= \frac{1}{|\calZ_k|}\sum_{(\boldx,\boldy)\in\calZ_k}
K(\boldx,\boldx_\ell)L(\boldy,\boldy_\ell).
\end{align*}

This procedure is repeated for $k = 1, \ldots, K$,
and its average $J^{(K\text{-CV})}$ is taken as
\begin{align*}
  J^{(K\text{-CV})} := \frac{1}{K}\sum_{k = 1}^K J_{\calZ_k}^{(K\text{-CV})}.
\end{align*}
Finally, we compute $J^{(K\text{-CV})}$ for all model candidates, and choose the one with minimum $J^{(K\text{-CV})}$.

\subsection{Dependence Maximization}
%
Based on the empirical estimate of SMI, the dependence maximization problem is given as
\begin{align*}
\max_{\boldPi}~~ & \widehat{\mathrm{SMI}}(Z(\boldPi)).
\end{align*}
We here provide a computationally efficient approximation algorithm based on \emph{dynamic programming} (DP) \cite{bellman52}. 

Let us rewrite the empirical SMI, Eq.\eqref{SMIhat}, as
\begin{align*}
\widehat{\mathrm{SMI}}(Z(\boldPi))&= \frac{1}{2m}\sum_{i = 1}^{n_{\mathrm x}}\sum_{j = 1}^{n_{\mathrm y}} \delta(\pi_i^{\mathrm x}, \pi_j^{\mathrm y}) r_{\widehat{\boldalpha}_{\boldPi}}(\boldx_i, \boldy_j) - \frac{1}{2},
\end{align*}
where 
\begin{eqnarray*}
\delta(\pi_i^{\mathrm x}, \pi_j^{\mathrm y}) = \left\{ \begin{array}{ll}
1 & \textnormal{if}~\boldx_{\pi_i^{\mathrm x}}~\textnormal{and}~\boldy_{\pi_j^{\mathrm y}}~\textnormal{are paired}, \\
0 & \textnormal{otherwise}. \\
\end{array} \right.
\end{eqnarray*}

Then, the solution is updated with the current $\boldPi^{\textnormal{old}}$ as
\begin{align}
\label{eq:DP-LSDTW}
\boldPi^{\textnormal{new}} = \argmax_{\boldPi} \sum_{i = 1}^{n_{\mathrm x}}\sum_{j = 1}^{n_{\mathrm y}} \delta(\pi_i^{\mathrm x}, \pi_j^{\mathrm y}) r_{\widehat{\boldalpha}_{\boldPi^{\textnormal{old}}}}(\boldx_i, \boldy_j).
\end{align}
This problem can be efficiently solved by DP with time complexity $O(n_{\mathrm x}n_{\mathrm y})$ (see Appendix). Note, however, that the solution to Eq.\eqref{eq:DP-LSDTW} does not always increase the empirical SMI, Eq.\eqref{SMIhat}; we update the alignment matrix $\boldPi$ only if the SMI score increases after the update. 

\section{Related Methods} \label{sec:related}
In this section, we review existing temporal alignment methods which are based on pairwise distance minimization (not dependence maximization) and point out their potential weaknesses.

\subsection{Dynamic Time Warping (DTW)}
The goal of \emph{dynamic time warping} (DTW) is, given two sequences of the \emph{same} dimensionality and the \emph{different} number of samples,
$\{\boldx_i\;|\; \boldx_i \in \mathbbR^{d}\}_{i=1}^{n_{\mathrm x}}$
and
$\{\boldy_j\;|\; \boldy_j \in \mathbbR^{d}\}_{j=1}^{n_{\mathrm y}}$,
to find an alignment such that the sum of pairwise distances between two sets is minimized \cite{saoke78,Rabiner:1993dq}:
\begin{align*}
\min_{\boldpi^{\mathrm x}, \boldpi^{\mathrm y}} & \sum_{i = 1}^m \| \boldx_{\pi^{\mathrm x}_i} - \boldy_{\pi^{\mathrm y}_i} \|^2,
\end{align*}
where $m$ is the number of indices needed to align the sequences. $\boldpi^{\mathrm x}$ and $\boldpi^{\mathrm y}$ need to satisfy the boundary, continuity, and monotonicity conditions (see Section \ref{sec:Problem}). The above DTW optimization problem
can be efficiently solved by DP with time complexity $O(n_{\mathrm x}n_{\mathrm y})$.

A potential weakness of DTW is that it cannot handle sequences with different dimensions such as image to audio alignment. Moreover, even when the dimensionality of sequences is the same, DTW may not be able to find a good alignment of sequences with different characteristics such as  sequences with different amplitudes. These drawbacks highly limit the applicability of DTW.
\subsection{Canonical Time Warping (CTW)}
\emph{Canonical time warping} (CTW) can align sequences with different dimensions in a common latent space \cite{NIPS2009_0760,CVPR:Feng:2012a}.

The CTW optimization problem is given as
\begin{align}
\label{eq:CTW}
\min_{\boldW_{\mathrm x}, \boldW_{\mathrm y},\boldV_{\mathrm x}, \boldV_{y}} & \| \boldV_{\mathrm x}^\top \boldX \boldW_{\mathrm x}^\top - \boldV_{\mathrm y}^\top \boldY \boldW_{\mathrm y}^\top\|^2_{\mathrm{Frob}},
\end{align}
where $\|\cdot\|_{\mathrm{Frob}}$ is the Frobenius norm, $\boldX = [\boldx_1, \ldots, \boldx_{n_{\mathrm x}}] \in \mathbbR^{d_{\mathrm x} \times n_{\mathrm x}}$, $\boldY = [\boldy_1, \ldots, \boldy_{n_{\mathrm y}}] \in \mathbbR^{d_{\mathrm y} \times n_{\mathrm y}}$, $\boldW_{\mathrm x} \in \{0,1\}^{m \times n_{\mathrm x}}$ and $\boldW_{\mathrm y} \in \{0,1\}^{m \times n_{\mathrm y}}$ are binary selection matrices that need to be estimated to align $\boldX$ and $\boldY$, and $\boldV_{\mathrm x} \in \mathbbR^{d_{\mathrm x} \times b}$ and $\boldV_{\mathrm y} \in \mathbbR^{d_{\mathrm y} \times b}$ ($b \leq \min(d_{\mathrm x}, d_{\mathrm y}))$ are linear projection matrices of $\boldx$ and $\boldy$ onto a common latent space, respectively. The above optimization problem can be efficiently solved by alternately solving CCA and DTW, where the alignment matrix obtained using DTW is usually used as initialization (initial alignment matrix).

However, since CTW finds a common latent space using CCA, it can only deal with linear and Gaussian temporal alignment problems. Thus, CTW cannot properly deal with multi-modal and non-Gaussian data. 
Another limitation of CTW is that
comparing the alignment quality over different model parameters
is not straightforward.
This is because, for different model parameters,
a common latent space found by CCA is generally different 
and thus the metric of the pairwise distance Eq.\eqref{eq:CTW} is also different.
For this reason, a systematic model selection method for
the regularization parameter, dimensionality of the common latent space,
and the initial alignment matrix has not been developed so far,
to the best of our knowledge.


\subsection{Dynamic Manifold Temporal Warping (DMTW)}
\emph{Dynamic manifold temporal warping} (DMTW) is a non-linear extension of CTW \cite{DBLP:conf/iccv/GongM11}.

The DMTW optimization problem is defined as
 \begin{align*}
\min_{\boldW_{\mathrm x}, \boldW_{\mathrm y},\calF_{\mathrm x}, \calF_{\mathrm y}} & \| \calF_{\mathrm x}(\boldX) \boldW_{\mathrm x}^\top - \calF_{\mathrm y}(\boldY) \boldW_{\mathrm y}^\top\|^2_{\mathrm{Frob}},
\end{align*}
where $\calF_{\mathrm x}: \mathbbR^{d_{\mathrm x} \times n_{\mathrm x}} \rightarrow \mathbbR^{b \times n_{\mathrm x}}$ and $\calF_{\mathrm y}: \mathbbR^{d_{\mathrm y} \times n_{\mathrm y}} \rightarrow \mathbbR^{b \times n_{\mathrm y}}$ are non-linear mapping functions that map $\boldx$ and $\boldy$ to a common latent subspace. 
DMTW first maps $\boldX$ and $\boldY$ to a one-dimensional smooth manifold (i.e., $b = 1$) by the \emph{tensor voting} method \cite{DBLP:journals/jmlr/MordohaiM10} and then align sequences on the manifold. 

DMTW highly depends on a specific non-linear transformations and requires the smooth manifold assumption. Thus, the usage of DMTW is limited to specific applications.
On the other hand, CTW and LSDTW do not require the latter strong assumption and thus can be useful for a broader range of applications.
This assertion will be experimentally validated in the next section.

\section{Experiments} \label{sec:experiments}
In this section, we experimentally evaluate our proposed LSDTW method
on synthetic and real-world \emph{Kinect} action recognition tasks.

\subsection{Setup}
In LSDTW, we use the Gaussian kernels:
\begin{align*}
K(\boldx, \boldx') = \exp \left(-\frac{\|\boldx - \boldx'\|^2}{2\sigma_{\mathrm{x}}^2} \right),~
L(\boldy, \boldy') = \exp \left(-\frac{\|\boldy - \boldy'\|^2}{2\sigma_{\mathrm{y}}^2} \right),
\end{align*}
where $\sigma_\mathrm{x}$, $\sigma_\mathrm{y}$, and $\lambda$ are chosen by 3-fold CV from 
\[
(\sigma_\mathrm{x}, \sigma_\mathrm{y}) = c\times(m_\mathrm{x}, m_\mathrm{y}),~c = 2^{-1/2}, 1.8^{-1/2}, \ldots, 0.2^{-1/2}, 
~~\lambda =10^{-1}, 10^{-2},
\]
and 
\[
m_\mathrm{x} = 2^{-1/2}\textnormal{median}(\{\|\boldx_i - \boldx_j\|\}_{i,j=1}^{n_{\mathrm x}}),~m_\mathrm{y} =  2^{-1/2}\textnormal{median}(\{\|\boldy_i - \boldy_j\|\}_{i,j=1}^{n_{\mathrm y}}).
\]

Due to non-convex nature of the objective, setting a good initial alignment is an important issue for LSDTW. 
Here, from the alignment obtained using CTW and the simple uniform initialization,
 \begin{align*}
\boldpi^{\mathrm x} &= [1, \lfloor 1 +  n_{\mathrm x}/m \rfloor, \lfloor 1+2n_{\mathrm x}/m \rfloor, \ldots, n_{\mathrm x}]^\top \in \mathbbR^{m \times 1}, \\
\boldpi^{\mathrm y} &= [1, \lfloor 1 + n_{\mathrm y}/m \rfloor, \lfloor 1+2n_{\mathrm y}/m \rfloor, \ldots, n_{\mathrm y}]^\top \in \mathbbR^{m \times 1}, 
\end{align*}
where $m = \min(n_{\mathrm x}, n_{\mathrm y})$,
we choose the one with the largest SMI score as the initial alignment for LSDTW. 

We compare the performance of LSDTW with DTW and CTW. For CTW, we choose the dimensionality of CCA to preserve 90\% of the total correlation, and we fix the regularization parameter at $0.01$. We use the alignment given by DTW as the initial alignment of CTW.

To evaluate the alignment results, we use the following standard alignment error \cite{CVPR:Feng:2012a}:
\begin{align*}
Error = \frac{\mathrm{dist}(\boldPi^\ast, \widehat{\boldPi}) 
+ \mathrm{dist}(\widehat{\boldPi},\boldPi^\ast)}{m^\ast + \widehat{m}},
~~~\mathrm{dist}(\boldPi_1, \boldPi_2) = \sum_{i = 1}^{m_1} \min(\{\|\boldpi_1^{(i)} - \boldpi_2^{(j)}\|\}_{j = 1}^{m_2}),
\end{align*}
where $\boldPi^\ast$ and $\widehat{\boldPi}$ are true and estimated alignment matrices and $\boldpi_1^{(i)}, \boldpi_2^{(j)} \in \mathbbR^{2\times1}$ are the $i$-th and $j$-th row of $\boldPi_1$ and $\boldPi_2$, respectively.

\begin{figure*}[t!]
\centering
\begin{minipage}[t]{0.315\linewidth}
\centering
  {\includegraphics[width=0.99\textwidth]{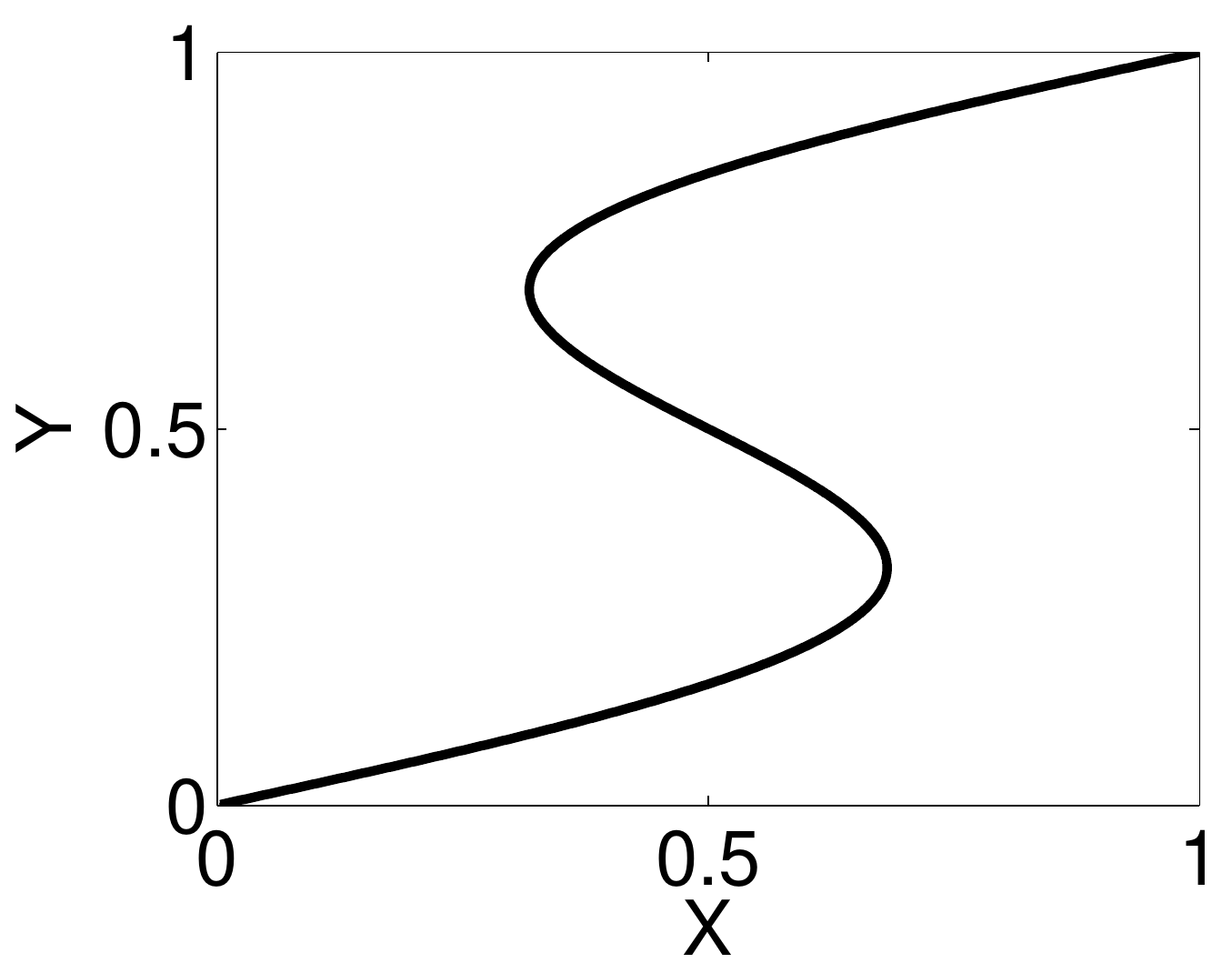}} \\ \vspace{-0.10cm}
(a-1) Multi-modal data.
\end{minipage}
\begin{minipage}[t]{0.325\linewidth}
\centering
  {\includegraphics[width=0.99\textwidth]{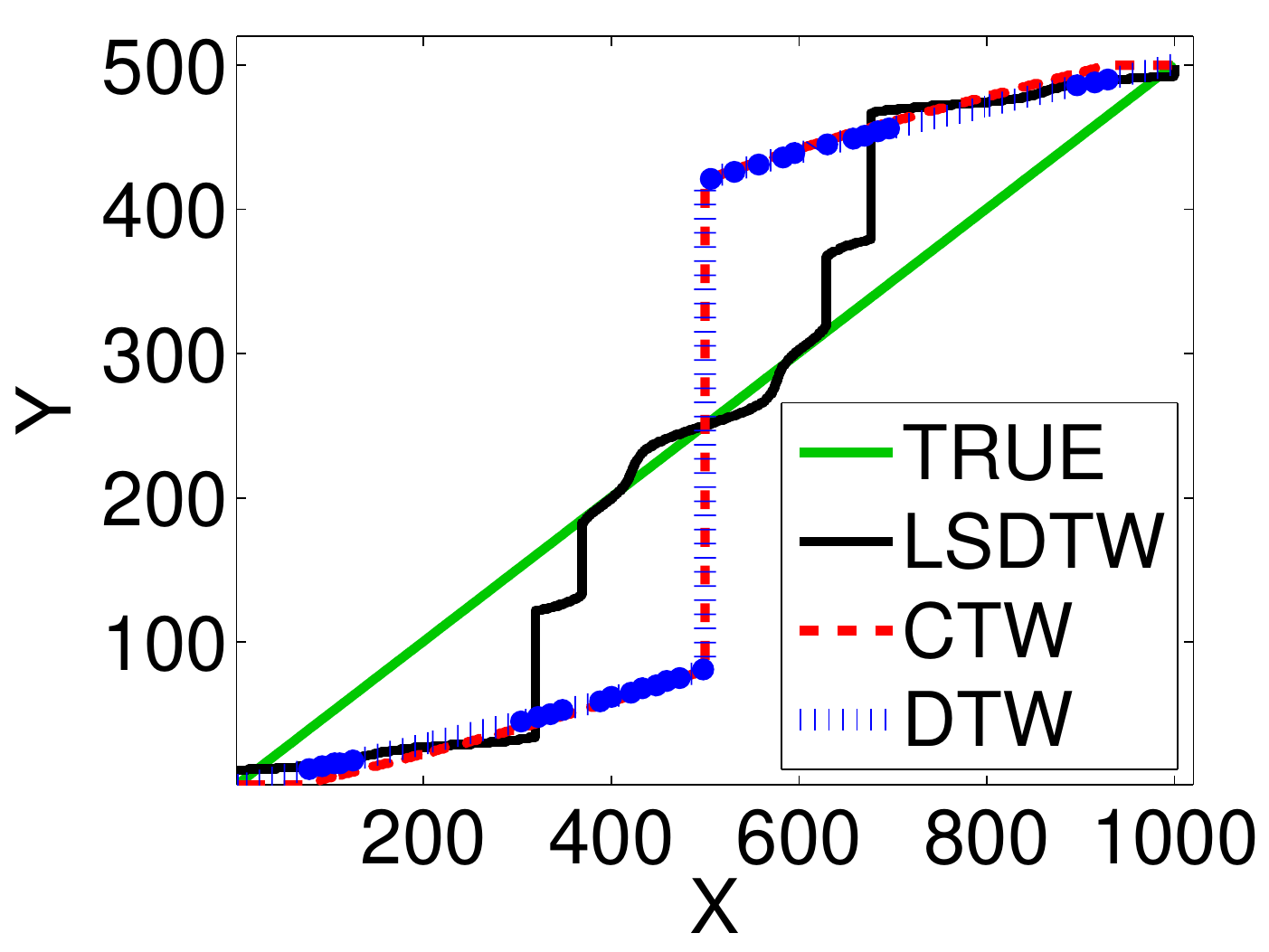}} \\ \vspace{-0.10cm}
(a-2) Alignment path.
\end{minipage}
\begin{minipage}[t]{0.31\linewidth}
\centering
  {\includegraphics[width=0.99\textwidth]{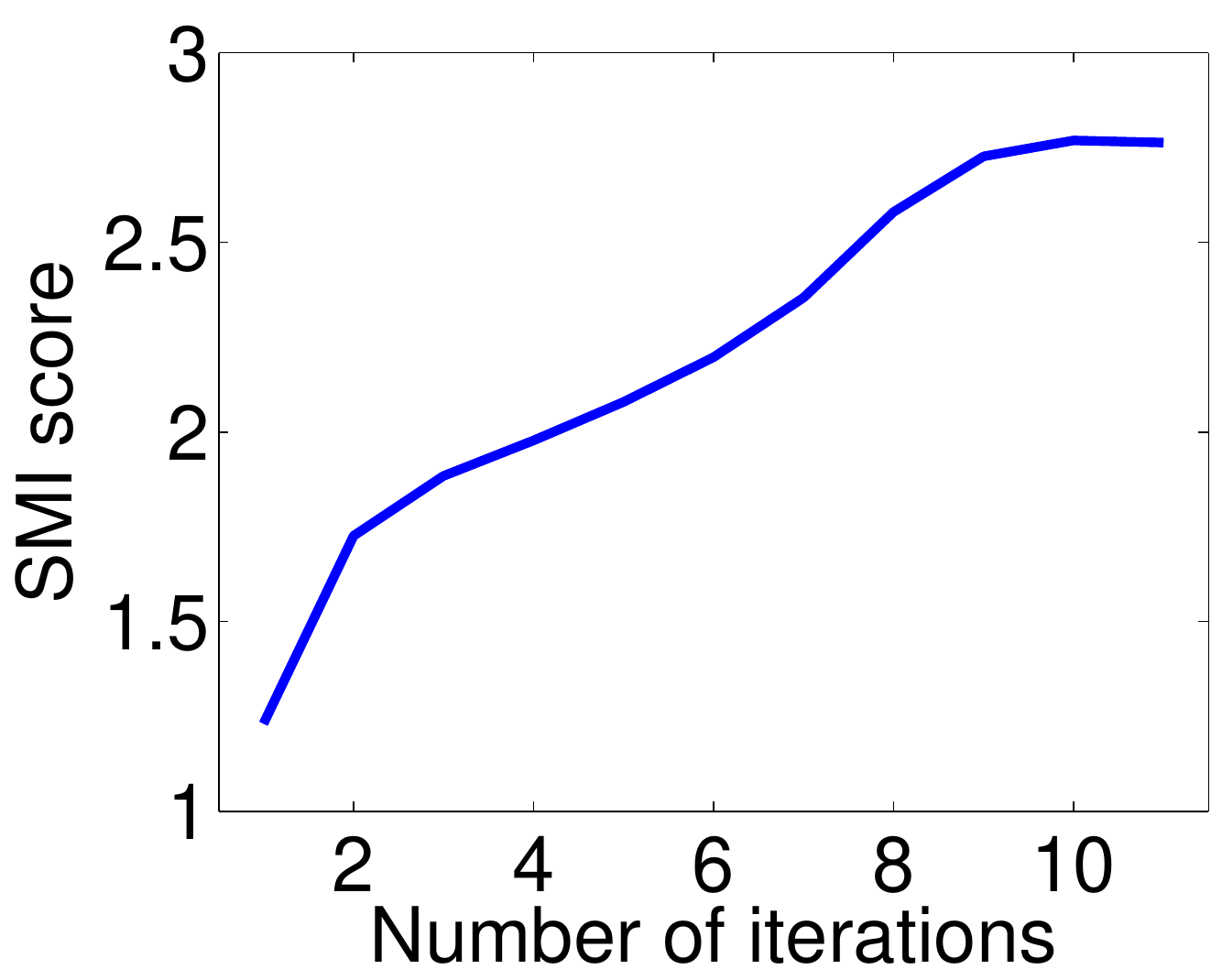}} \\ \vspace{-0.10cm}
(a-3) SMI score.
\end{minipage}\\[3mm]
\begin{minipage}[t]{0.305\linewidth}
\centering
  {\includegraphics[width=0.99\textwidth]{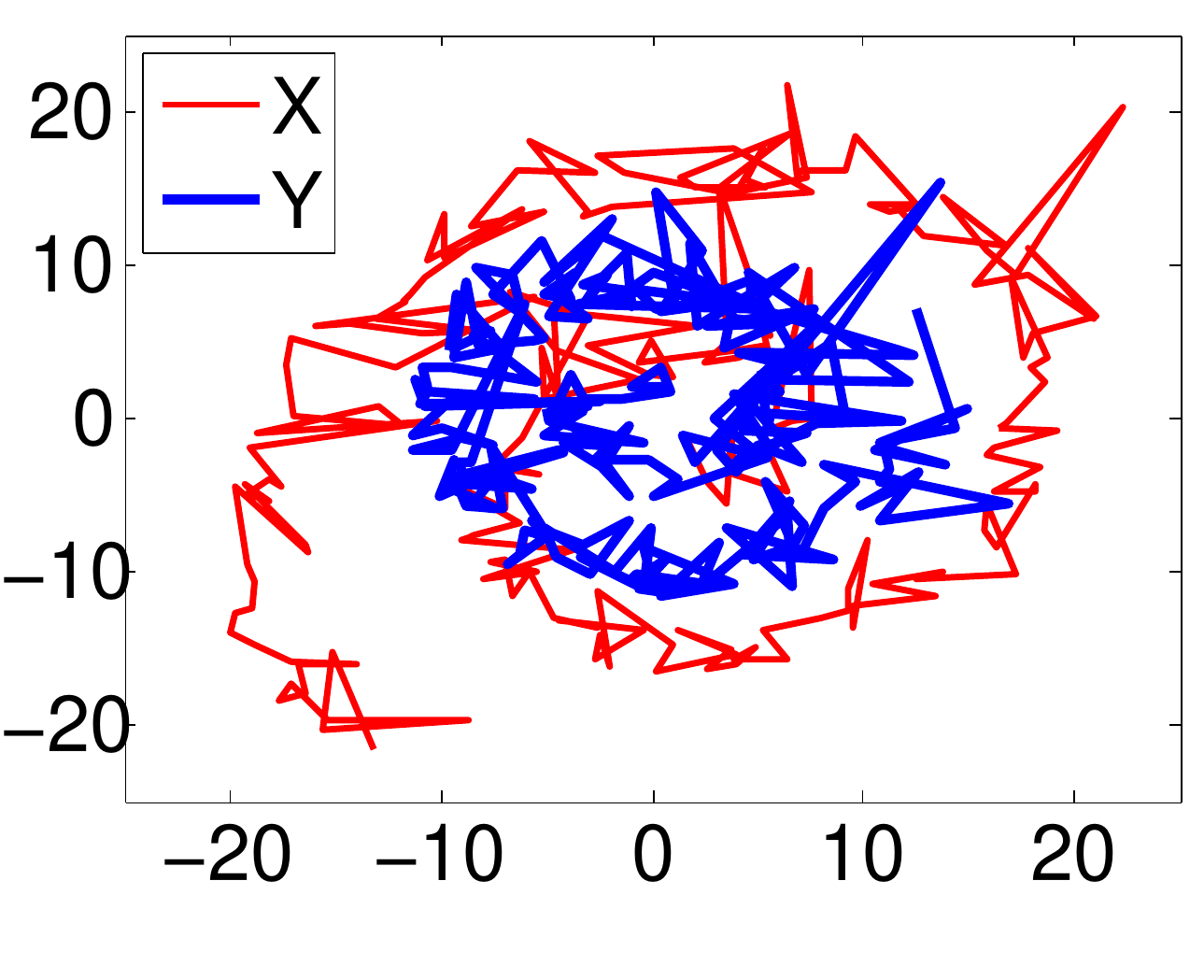}} \\ \vspace{-0.10cm}
(b-1) Non-Gaussian data.
\end{minipage}
\begin{minipage}[t]{0.325\linewidth}
\centering
  {\includegraphics[width=0.99\textwidth]{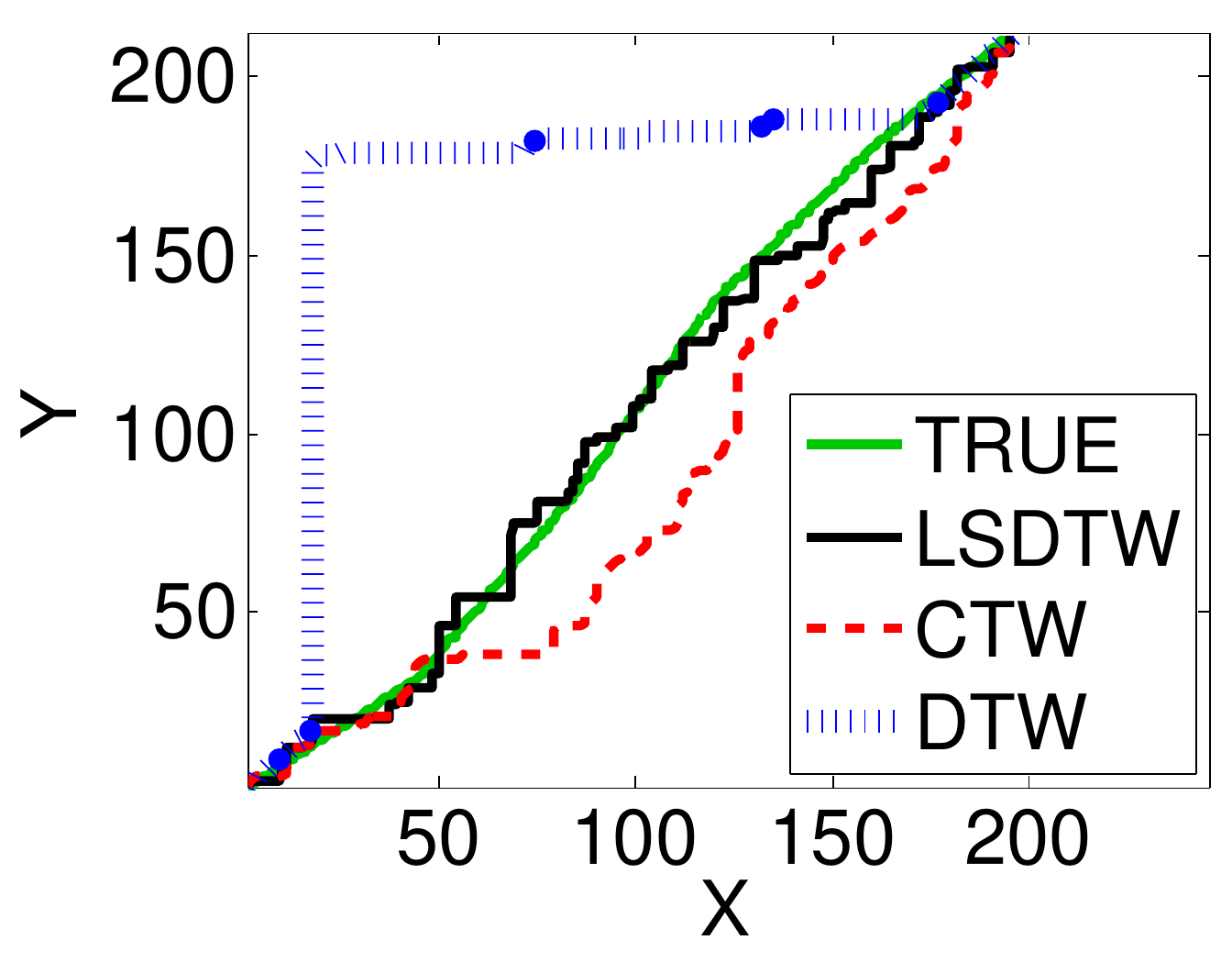}} \\ \vspace{-0.10cm}
(b-2) Alignment path.
\end{minipage}
\begin{minipage}[t]{0.305\linewidth}
\centering
  {\includegraphics[width=0.99\textwidth]{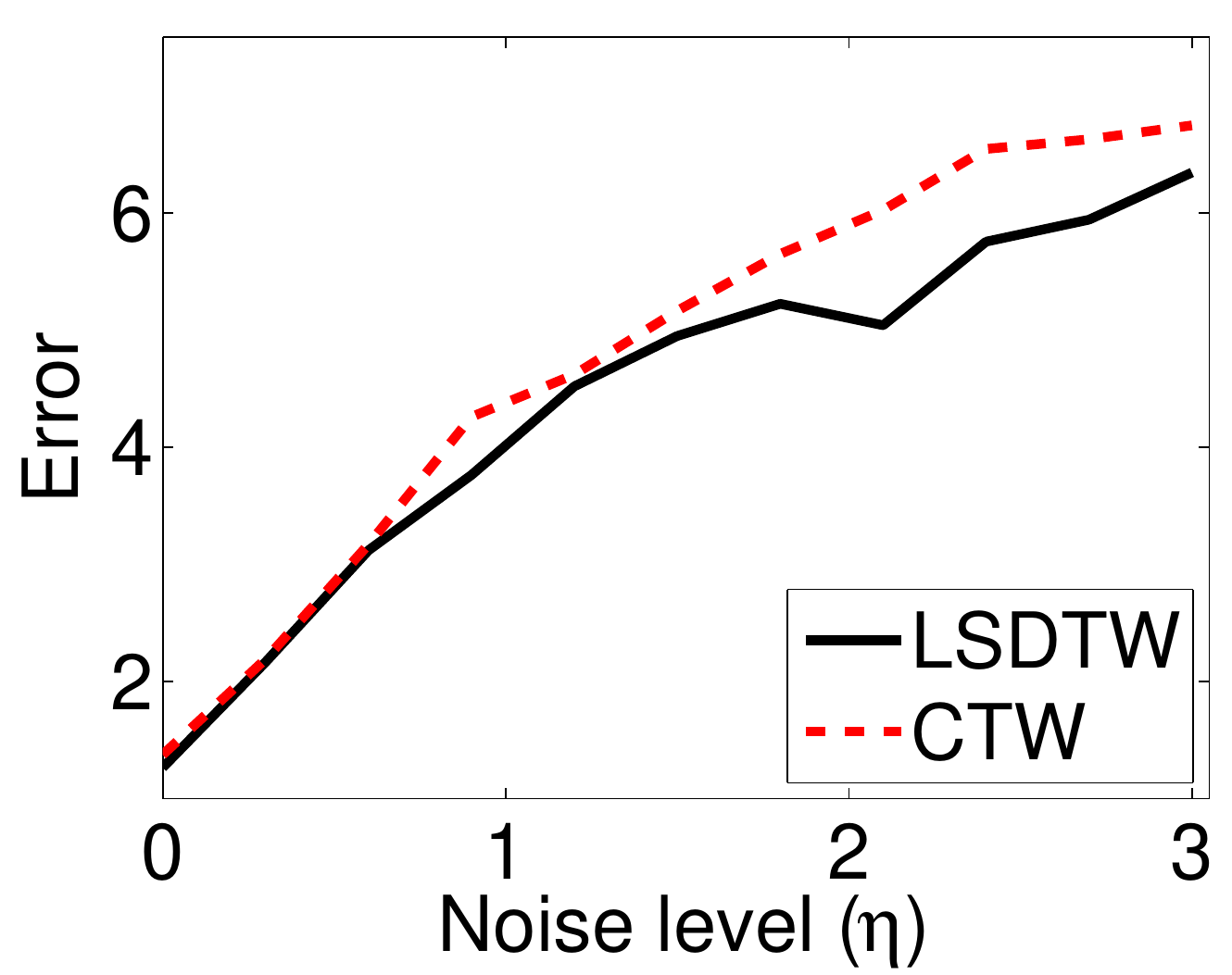}} \\ \vspace{-0.10cm}
(b-3) Mean alignment error.
\end{minipage}
\caption{Results of synthetic experiments. (a-1) Multi-modal data. (a-2) Alignment paths. (a-3) SMI score as a function of the number of iterations. (b-1) A synthetic dataset with an additive exponential noise ($\eta = 2.4$). (b-2) Alignment paths. Here, the alignment error of LSDTW, CTW, and DTW are 5.93, 26.28, and 87.69, respectively. (b-3) The mean alignment error over 100 runs as functions of  noise level $\eta$. }
    \label{fig:result_toy}
    \vspace{-0.1in}
\end{figure*}

\subsection{Synthetic Dataset}
First, we illustrate the behavior of the proposed LSDTW method for non-linear and non-Gaussian data using synthetic datasets.

{\bf Non-linear (multi-modal) data\footnote{A distribution of a multi-modal data has two or more modes.}:}
\begin{align*}
x_i &= z_i + 0.4\sin(2\pi z_i),~i = 1,\ldots,1000, \\
y_j &= z_{(j -1)\times 2 + 1},~j=1,\ldots,500,
\end{align*}
where $z_i = i/1000$
(see Figure~\ref{fig:result_toy}-(a-1)).

Figure~\ref{fig:result_toy}-(a-2) shows the alignment path obtained by LSDTW, CTW, and DTW, respectively. In this experiment, we initialize LSDTW and CTW with the true alignment matrix and check whether those methods perform well. As can be observed, LSDTW can find a better alignment in the middle region (i.e., a multi-modal region) than DTW and CTW. 
This shows that LSDTW objective is much better than alternatives when it comes to multi-modal data.  
 Figure~\ref{fig:result_toy}-(a-3) depicts the SMI score with respect to the number of iterations in LSDTW, showing that LSDTW converges in 10 iterations.

{\bf Non-Gaussian data:}
\begin{align*}
\boldX = \boldU_{\mathrm x}^\top \boldZ \boldM_{\mathrm x}^\top + \eta \boldE_{\mathrm x},
~~
\boldY = \boldU_{\mathrm y}^\top \boldZ \boldM_{\mathrm y}^\top + \eta \boldE_{\mathrm y},
\end{align*}
where $\boldU_{\mathrm x}, \boldU_{\mathrm y} \in \mathbbR^{2 \times 2}$ are randomly generated affine transformation matrices, $\boldZ \in \mathbbR^{2 \times m}$ is a trajectory in two dimensions, $\boldM_{\mathrm x} \in \mathbbR^{n_{\mathrm x} \times m}$ and $\boldM_{\mathrm y} \in \mathbbR^{n_{\mathrm x} \times m}$ are randomly generated matrices for time warping, $\boldE_{\mathrm x} \in \mathbbR^{2 \times n_{\mathrm x}}$ and $\boldE_{\mathrm y} \in \mathbbR^{2 \times n_{\mathrm y}}$ are randomly generated additive exponential noise with rate parameter 1 (and its mean is adjusted to be zero), and $\eta = \{0, 0.3, 0.6,\ldots, 3.0\}$ is the noise level. Note that larger noise level $\eta$ means stronger non-Gaussianity in the data.

Figures~\ref{fig:result_toy}-(b-1) and (b-2) show an example of synthetic data with additive exponential noise ($\eta = 2.4$) and corresponding alignment paths obtained by LSDTW, CTW, and DTW. As can be seen, only the proposed method can find a good alignment.
 Figure~\ref{fig:result_toy}-(b-3) shows the mean alignment error over 100 runs,
from which we can confirm that
the proposed method tends to outperform the existing methods for larger noise levels.

\subsection{Real-world Kinect Action Recognition Data}
Next, we evaluate the proposed LSDTW method on the publicly available \emph{Kinect} action recognition dataset\footnote{\url{www.cs.ucf.edu/~smasood/datasets/UCFKinect.zip}} \cite{DBLP:conf/iccvw/MasoodNKZT11}. This dataset consists of the human skeleton data (15 joints) obtained using a \emph{Kinect} sensor, and there are 16 subjects and 16 actions with 5 runs. Instead of using the raw skeleton data, we here use the 105-dimensional feature vector, where each element of the feature vector is the Euclidean distance between joint pairs. 

In evaluation, we carry out \emph{unsupervised} action recognition experiments and evaluate the performance of alignment methods by classification accuracy. More specifically, we first divide the action recognition dataset into two disjoint subsets: 8 subjects (\#1-\#8) with all actions for testing (in total 640 sequences) and the remaining subjects (\#9-\#16) with all actions for ``training" database (in total 640 sequences). Then, we retrieve $N=10$ similar actions for each test action from the database by DTW, CTW, and LSDTW. Here, we use the pairwise Euclidean distance based on an estimated alignment to measure the similarity between sequences. Finally, if there is at least one correct action in the retrieved sequences, we regard the action to be correctly retrieved.

\begin{figure*}[t!]
\centering
\begin{minipage}[t]{0.325\linewidth}
\centering
  {\includegraphics[width=0.99\textwidth]{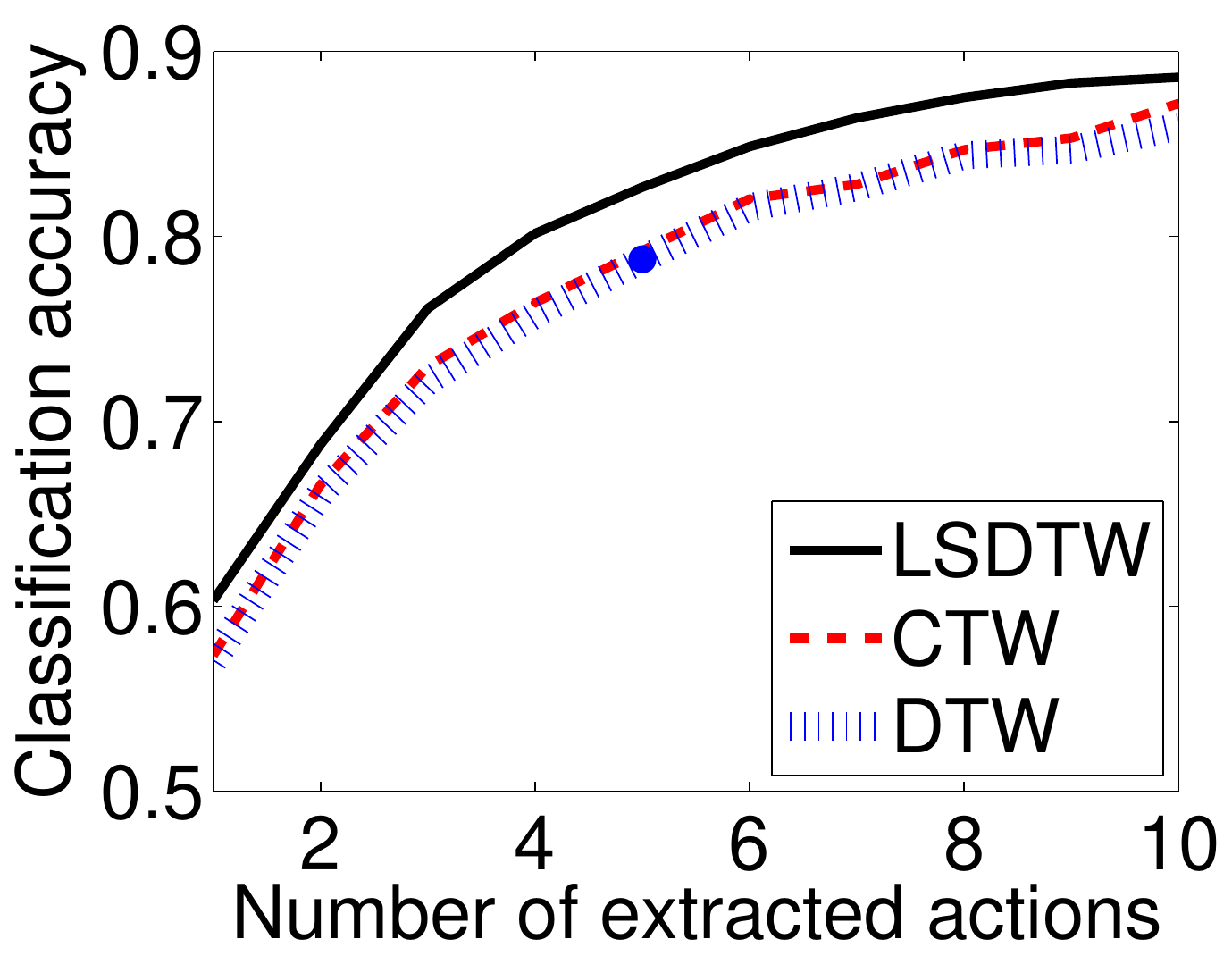}} \\ \vspace{-0.10cm}
(a) 
\end{minipage}
\begin{minipage}[t]{0.325\linewidth}
\centering
  {\includegraphics[width=0.99\textwidth]{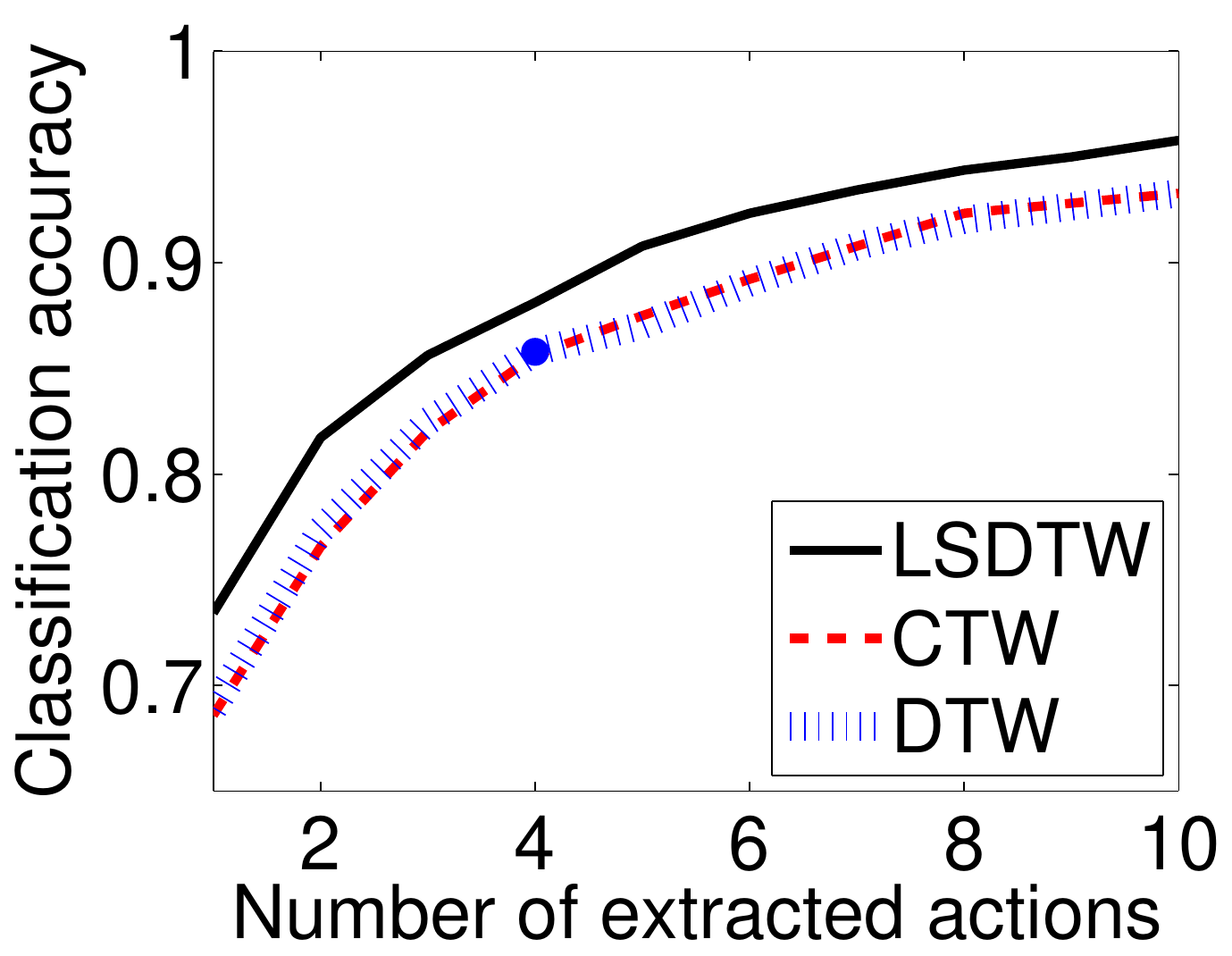}} \\ \vspace{-0.10cm}
(b) 
\end{minipage}
\begin{minipage}[t]{0.325\linewidth}
\centering
  {\includegraphics[width=0.99\textwidth]{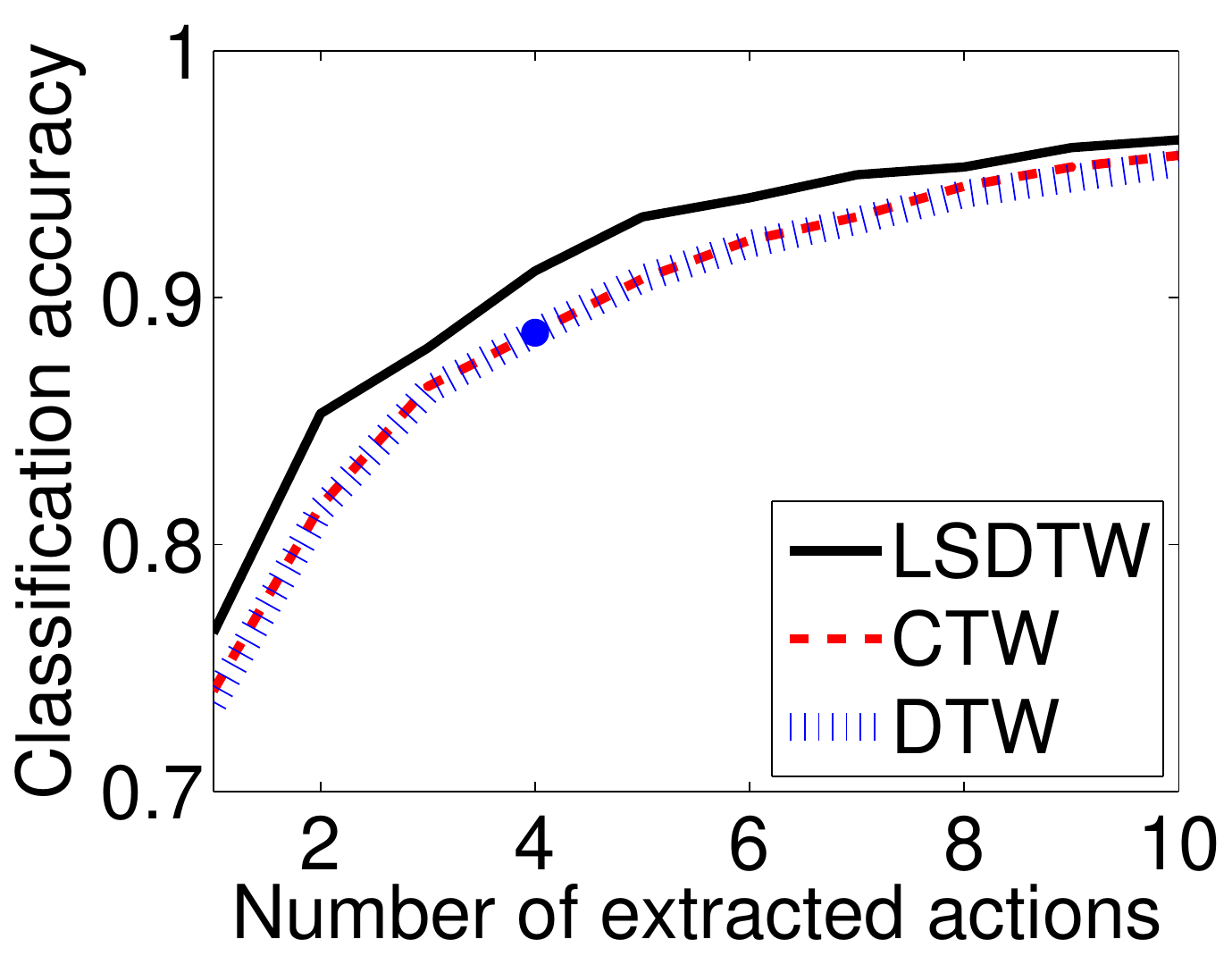}} \\ \vspace{-0.10cm}
(c) 
\end{minipage}
\caption{Mean classification accuracy with respect to the number of retrieved actions. (a) Only sequences of subject$\#9$ are used as database. (b) Sequences of Subjects$\#9-12$ are used as database. (c) Sequences of all subjects ($\#9-16$) are used as database. }
    \label{fig:result_kinect}
\end{figure*}

Figure~\ref{fig:result_kinect} shows the mean classification accuracy as functions of the number of retrieved actions, $N$, where three different database sizes are tested.
The graphs clearly show that LSDTW compares favorably with existing methods in terms of classification accuracy. 

\section{Conclusions}
In this paper, we proposed a novel temporal alignment framework called the \emph{dependence maximization temporal alignment} (DMTA) and developed a DMTA method called the \emph{least-squares dynamic time warping} (LSDTW). LSDTW adopts \emph{squared-loss mutual information} as a dependence measure, which is efficiently estimated by the method of \emph{least-squares mutual information}. Notable advantages of LSDTW are that it can naturally deal with non-linear and non-Gaussian sequences and it can optimize model parameters such as the Gaussian kernel width and the regularization parameter by cross-validation. We applied the proposed method on the \emph{Kinect} action recognition task, and experimentally showed that LSDTW is the promising alternative to the compared methods.

\section*{Acknowledgments}
We thank Prof.~Fernando De la Torre, Mr.~Feng Zhou, and Dr. Akisato Kimura for their valuable comments. MY acknowledges the JST PRESTO and PLIP programs, and MS acknowledges the JST PRESTO program and AOARD for financial support.

\bibliography{main}

\begin{thebibliography}{10}
\bibitem{PhMag:Pearson:1900}
K.~Pearson.
\newblock On the criterion that a given system of deviations from the probable
  in the case of a correlated system of variables is such that it can be
  reasonably supposed to have arisen from random sampling.
\newblock {\em Philosophical Magazine}, 50:157--175, 1900.
\end{thebibliography}
\bibliographystyle{unsrt}

\section*{Appendix}
Given the empirical estimate of SMI computed at the dependence estimation step (Sect. 2.2.2), the depence maximization problem is given as 
\begin{align}
\begin{split}
\max_{\boldPi}~~  \widehat{\mathrm{SMI}}(Z(\boldPi))  &~~=~ \max_{\boldPi}~~  \frac{1}{2m} \sum_{i = 1}^{m} r_{\widehat{\boldalpha}_{\boldPi_{\mathrm{old}}}}(\boldx_{\pix_i}, \boldy_{\piy_i}) - \frac{1}{2} \\
&\overset{\text{Eq. (3)}}= \max_{\boldPi}~~  \frac{1}{2 m} \sum_{i = 1}^{m} \sum_{\ell = 1}^{m_{\mathrm{old}}} {\widehat{\alpha}_{\ell}} K(\boldx_{\pix_i},\boldx_{{\pix_\ell}_{\mathrm{old}}} ) L(\boldy_{\piy_i},\boldy_{{\piy_\ell}_{\mathrm{old}}} ) - \frac{1}{2}\\
& ~~\approx~  \frac{1}{2( n_{\mathrm x} + n_{\mathrm y})} \max_{\boldPi}~ \sum_{i = 1}^{m} \sum_{\ell = 1}^{m_{\mathrm{old}}} {\widehat{\alpha}_{\ell}} K(\boldx_{\pix_i},\boldx_{{\pix_\ell}_{\mathrm{old}}} ) L(\boldy_{\piy_i},\boldy_{{\piy_\ell}_{\mathrm{old}}} ) - \frac{1}{2}.
\end{split}
\end{align}
Based on the constraints on the alignment functions $\boldPi$ described in Sect. 2.1, this optimal alignement can be computed by dynamic programming (DP) \cite{bellman52}. In order to verify this, we define the prefix sequences 
$\boldX_n := \{\boldx_i\;|\; \boldx_i \in \mathbbR^{d_{\mathrm x}}\}_{i=1}^{n} \text{ and }
\boldY_{n'} := \{\boldy_j\;|\; \boldy_j \in \mathbbR^{d_{\mathrm y}}\}_{j=1}^{n'} $, with $n \leq n_{\mathrm x} \text{ and } n' \leq n_{\mathrm y}$, and set $A(n,n') :=  \widehat{\mathrm{SMI}}(\boldX_{n}, \boldY_{n'})$ denoting the optimal SMI for the aligned prefix sequences $\boldX_n$ and $\boldY_{n'}$.

Following the boundary conditions of the alignement functions, we have:
\begin{equation}
 A(1,1) = r_{\widehat{\boldalpha}_{\boldPi_{\mathrm{old}}}}(\boldx_{1}, \boldy_{1}).
\end{equation}
Based on the continuity and monotonicity conditions, the DP-equation is given as
\begin{equation}
A(n,n') = \max \{  A(n-1, n'-1), A(n-1, n'), A(n,n'-1) \}+ r_{\widehat{\boldalpha}_{\boldPi_{\mathrm{old}}}}(\boldx_{n}, \boldy_{n'}),
\end{equation}
\noindent for $1 < n \leq n_{\mathrm x}$ and $1< n' \leq n_{\mathrm y}$. Therefore, the optimal  
$\widehat{\mathrm{SMI}}(Z(\boldPi)) =  \frac{1}{2( n_{\mathrm x} + n_{\mathrm y})} A(n_{\mathrm x} ,n_{\mathrm y}) - \frac{1}{2}$ can be computed in $ O(n_{\mathrm x} n_{\mathrm y})$. Given the accumulated cost matrix $A$, we can compute the optimal alignment  $\boldPi$ using backtracking.

\end{document}